\documentclass[10pt, conference, compsocconf]{IEEEtran}
\usepackage{times}
\usepackage{epsfig}
\usepackage{graphicx}
\usepackage{amsmath}
\usepackage{amssymb}
\usepackage{subfig}
\usepackage[utf8]{inputenc}
\begin{document}
%
\title{Sign Language Fingerspelling Classification from Depth and Color Images using a Deep Belief Network}

\author{\IEEEauthorblockN{Lucas Rioux-Maldague \& Philippe Giguère}
\IEEEauthorblockA{Department of Computer Science\\
Laval University\\
Quebec City, QC\\
Email: lucas.rioux-maldague.1@ulaval.ca, philippe.giguere@ift.ulaval.ca}
}


%


\maketitle

\begin{abstract}  
  Automatic sign language recognition is an open problem that has received a lot of attention recently, not only because of its usefulness to signers, but also due to the numerous applications a sign classifier can have. In this article, we present a new feature extraction technique for hand pose recognition using depth and intensity images captured from a Microsoft Kinect\texttrademark sensor. We applied our technique to American Sign Language fingerspelling classification using a Deep Belief Network, for which our feature extraction technique is tailored. We evaluated our results on a multi-user data set with two scenarios: one with all known users and one with an unseen user. We achieved 99~\% recall and precision on the first, and 77~\% recall and 79~\% precision on the second. Our method is also capable of real-time sign classification and is adaptive to any environment or lightning intensity.

\end{abstract}


%
\IEEEpeerreviewmaketitle

\section{Introduction}

\label{sec_intro}

Automated recognition of hand signals has many applications in computer science. It might facilitate the interaction between humans and computers in many situations, especially for people with disabilities. One interesting area of focus is the recognition of sign languages. This way of communication is widely used among Deaf communities. Therefore, a system capable of understanding it would be convenient for them, just like automated speech recognition is useful to people using spoken languages. Also, such a system could be the base for numerous other applications and ways of interacting with computers as an alternative to traditional input devices such as mouse and keyboard.

However, sign languages usually include non-manual signs such as facial expressions in addition to thousands of manual gestures and poses. Additionnally, some of these signs can be sequential (like spoken languages) or parallel \cite{aran2008vision}. This large variety of signs adds to the already complex task of finding a body and hands in a  dynamic environment. 

\begin{figure}
 \centering
 \includegraphics[width=0.5\textwidth]{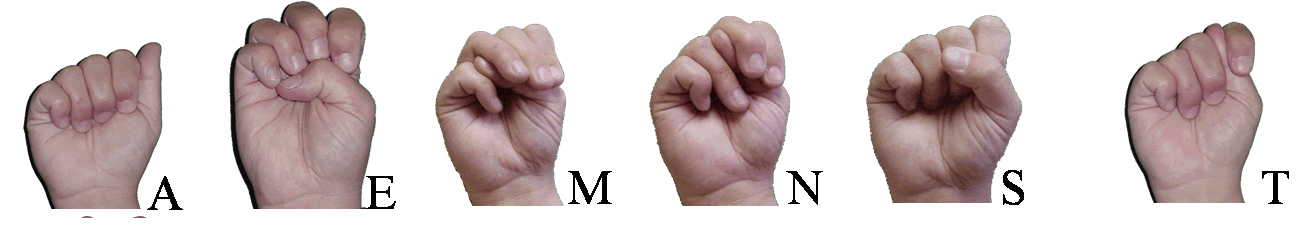}
 \caption{These six gestures only vary by the position of the thumb, making them the most frequently confused.}
 \label{fig_SimilarGestures}
\end{figure}

\begin{figure}
 \centering
 \subfloat{\includegraphics[width=0.08\textwidth] {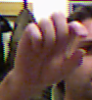}} \hspace{2px}
 \subfloat{\includegraphics[width=0.08\textwidth] {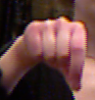}} \hspace{2px}
 \subfloat{\includegraphics[width=0.08\textwidth] {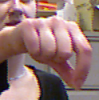}} \hspace{2px}
 \subfloat{\includegraphics[width=0.08\textwidth] {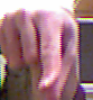}} \hspace{2px}
 \subfloat{\includegraphics[width=0.08\textwidth] {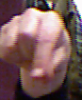}}
 \caption{Illustration of the difference between the five different users for the same sign (Q).}
 \label{fig_DifferentSigns}
\end{figure}

For this reason, research at the moment focuses mostly on specific, easier recognition tasks. One such interesting task is the recognition of American Sign Language (ASL) fingerspelling, which is a way of spelling words for which the sign is unknown or non existent in the ASL language. Although ASL does not share much with the English language syntactically, ASL fingerspelling uses the same 26 letter alphabet as written English to manually spell out words. Fingerspelling constitutes a significant portion of ASL 
exchanges, accounting for 12-35~\% of ASL communication \cite{padden2003alphabet}.  Fingerspelling is performed with a single hand and is composed of 24 static and two dynamic hand poses, some of which appear very similar. The two dynamic poses are J and Z, and simply involve drawing the letter with either the little finger (J) or index finger (Z). The task of classifying fingerspelling, even as a smaller subset of ASL, is still very challenging due to the similarity between poses (see Fig. \ref{fig_SimilarGestures}). Also, the hand pose can vary greatly between different users, due to different hand shapes and sizes, knowledge of ASL, previous signs, arm orientation and others. The data set we use (from \cite{pugeault2011spelling}) is challenging in that aspect, providing various orientations over five very different users (Fig. \ref{fig_DifferentSigns}). The system we propose is capable of on-the-fly classification of the 24 static signs, using inputs provided by a traditional intensity camera, in addition 
to a depth camera.

\begin{figure}
 \centering
 \includegraphics[width=0.5\textwidth]{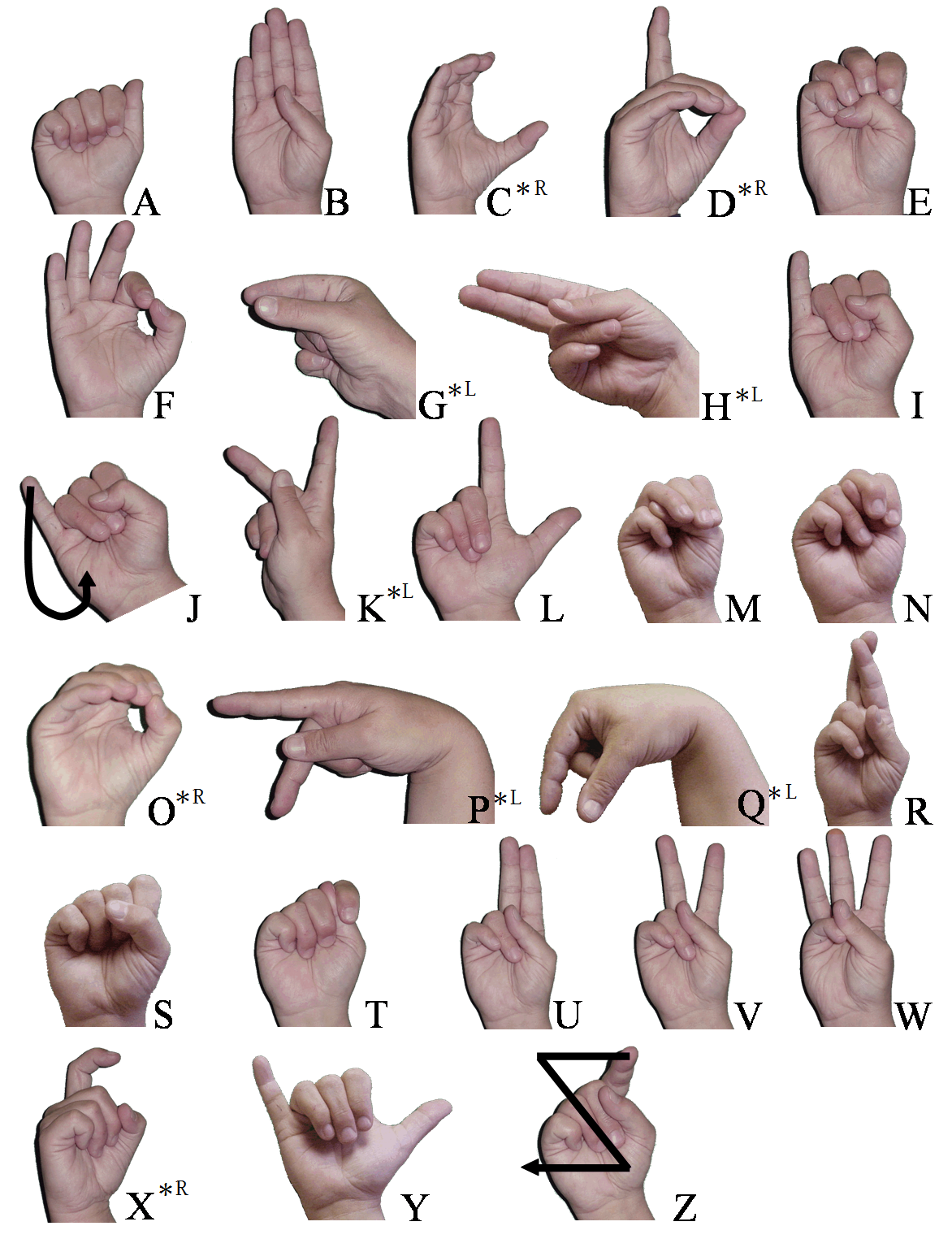}
 \caption{The American Manual Alphabet. Signs with a * are pictured from the side. Reproduced from \cite{ManuelAlphabet}}
 \label{fig_ManuelAlphabet}
\end{figure}

\section{Related Work}
\label{sec_relatedwork}

Hand pose recognition has already been the subject of much work. We divide this work in three sections: work on hand pose recognition using only regular intensity images, work using only depth images, and work using a mixed intensity-depth system.

\paragraph{Intensity Input Only}
Ong and Bowden~(2004) introduced a tree structure of boosted cascades that can recognize a hand and classify it in different shapes \cite{ong2004boosted}. They achieved an average error of 2.6~\% on their test set. However, their method is time-consuming and their classifier is built on automatically chosen shapes. Instead of classifying images of signs which are already categorized in a number of different categories, they use an algorithm to automatically cluster their training set in 300 clusters, each cluster containing similar images. If this was applied to ASL, it would cluster signs such as M and N together due to their similarity, thus eliminating the confusion problem. After this clustering, they trained their cascade on each cluster. Such a method does not allow for pre-determined hand shapes and, as such, is not well suited for ASL fingerspelling. 
Fang et al.~(2007) used a scale-space feature detection system and achieved 98.8~\% accuracy on six easily distinguishable signs \cite{fang2007real}. However, their accuracy drops to 89 \% when used on cluttered backgrounds. Liwicki and Everingham~(2009) attempted British Sign Language~(BSL) fingerspelling and achieved 84.1 \% accuracy using a Histogram of Oriented Gradients method \cite{liwicki2009automatic}. However, their method cannot be truly compared to ASL fingerspelling since BSL is two-handed, which introduces more variation between signs.

Of the work directly related to ASL fingerspelling, Isaacs and Foo~(2004) used a wavelet feature detector on regular intensity images combined with a two layer Multilayer perceptron (MLP) for the 24 static ASL letters \cite{isaacs2004hand}. They achieved 99.9~\% accuracy, but make no mention of the number of users and the existence of a training and testing set.

\paragraph{Depth Input Only}
Uebersax et al.~(2011) introduced the use of depth images for ASL fingerspelling using a Time-of-Flight (ToF) camera and an ANMM method for the 26 dynamic hand poses \cite{uebersax2011real}. On a seven user data set containing at least 50 images per user per letter, they achieved 76~\% average recognition rate.
 
\paragraph{Mixed Intensity-Depth Inputs}
Van Den Berg~(2011) introduced the use of a mixed intensity and depth method for hand-segmentation and gesture recognition \cite{van2011combining}. They reduced the dimensionality of the data using Average Neighborhood Margin Maximization (ANMM), then used a Nearest Neighborhood (NN) classifier. On 350 samples of six very different signs, they achieved almost perfect accuracy, but their signs very different from each other making it an easy task. This great variation between signs make the additional depth data superfluous, with accuracy varying by less than 0.5~\% between the results using depth data and the results using only intensity data.
Doliotis et al.~(2011) used a mixed depth and intensity classifier using a Kinect sensor \cite{doliotis2011comparing}. They applied their technique on the 10 digits of the Palm Graffiti alphabet. These are 10 dynamic gestures defined by a movement of the finger in the air. They used a 10 user data set along with a Nearest Neighbor classification framework. They achieve 95~\% accuracy, but their signs are very different from each other, and their technique only works for a small number of classes.
Pugeault and Bowden~(2011) introduced a classifier based on Gabor filters and trained using multi-class random forests. They achieved 54~\% recall and 75~\% precision on a data set of over 60000 examples recorded from five different users \cite{pugeault2011spelling}. They also used a mixed intensity-depth camera (Kinect) similarly to Uebersax et al. Finally, Kim, Livescu and Shakjnarovich~(2012) use Scale-invariant feature transform (SIFT) features, a mixed MLP - Hidden Markov Model (HMM) classifier and a dictionary prior for the 26 letters in ASL fingerspelling \cite{kim2012american}. They achieve a very low letter error rate of 9~\%, but they are helped by a prior on letter probability distribution and a previous letter bigram prior which boosts their accuracy. Their data set also only contains images from two users, which is much easier to learn than a data set containing more users.\\

Our system makes use of both intensity and depth images captured from a Kinect sensor. We train our classifier on a data set featuring five different users, some of whom making the signs in a very specific manner. Our work differs from previous attempts in many ways. First, we use the depth images as a mask on the intensity images to segment the hand from the background in order to achieve background independence. Second, we introduce a new feature extraction technique for the depth image. These new features capture the most important differences between each sign and are well tailored for Deep Belief Networks (DBN). Third, we use a DBN for our classifier, which is a deep multi-layer perceptron capable of learning interesting patterns in the data. Research on DBNs is currently very active, as it is considered a very promising technology. Finally, although the task of ASL fingerspelling classification on data provided by multiple users is challenging, our system achieves excellent results, surpassing previous 
results for our most difficult scenario.

\section{Hand Pose Classification}

In order to classify the different hand poses, we use two different input methods: regular intensity images and depth images. Depth images are matrices of the depth of the environment, where the value of each pixel corresponds to the depth of that pixel relative to the camera. Thus, using both intensity and depth, an accurate 3D perspective of an object can be created. Using depth data is becoming more common due to the increased number of 3D sensors available in the market. For our experiment, we used a Microsoft Kinect sensor. This device has a 640x480 image resolution for both intensity and depth images, with a depth range between two and eight meters.

In this article, we will only focus on the task of classifying the hand pose using hand images. Those images were obtained by tracking the hands using readily available functions in the OpenNI+NITE framework \cite{pugeault2011spelling}. They have a size of about 100x100 pixels and are framed around the hand, with the background still present. From these images, we extract features that are subsequently fed to a Deep Belief Network (DBN), which classifies the hand pose as one of 24 possible letters.

\subsection{Feature Extraction}
\label{sec_features}

The value of the pixels in the depth image extracted from the Kinect have a direct relation in mm to real world coordinates, with one unit representing one millimeter. A pre-processing step is done on the images before extracting the features. First, the background of the depth image is eliminated using a threshold $t$ on the maximum hand depth. To do this, since the hand is always the closest object to the camera in the extracted frames, its depth $d$ is computed by finding the smallest non-zero value in the image (zero values are errors). Then, every pixel with a distance greater than $t + d$ is set to 0. After this, $d-1$ is subtracted from the value of every non-zero pixel of the depth image so that the closest pixel is always 1. This creates depth independence between all images. 

From this filtered depth image, a binary mask is created and used directly on the intensity image, after compensating for the different location of the camera by a scaling and a translation. The mask takes value 0 for the background and 1 for the hand pixels (pixels having value greater than 0). Since this is a binary mask, it only takes an element-wise multiplication between the mask and the intensity image to segment the hand from the background. After this, we resize both depth and intensity images to 128x128 and center the content using the bounding box technique.

After this pre-processing, features are extracted. On the intensity side, the image is de-interlaced by keeping every other line and resizing from 128x64 to 64x64. Artifacts of interlacing can be seen in Fig. \ref{fig_DifferentSigns}, in the form of phantom horizontal edges. Then, the image's intensity histogram is equalized and all the pixels are normalized to a $[0,1]$ interval. The result is a 64x64 image, which is unrolled as a 1x4096 vector $f_{intensity}$.

\begin{figure}
 \centering
 \includegraphics[width=0.3\textwidth]{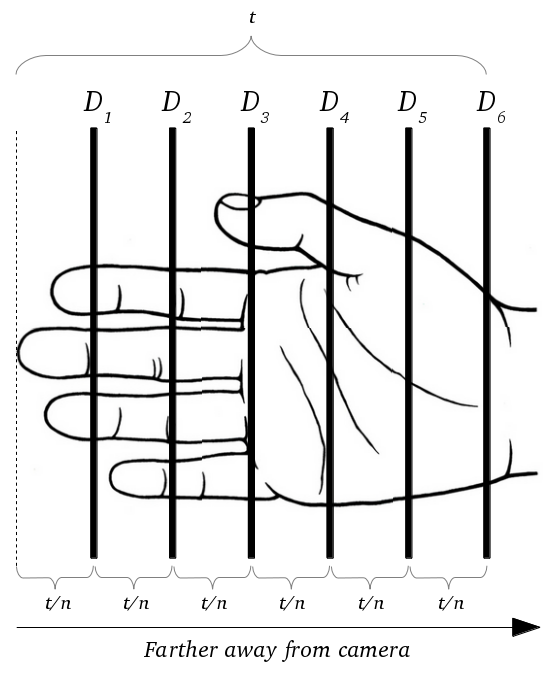}
 \caption{Drawing of the layers used to decompose the hand. Each binary image $D_l$ represents the parts of the hand to the left of the layer.}
 \label{img_layers}
\end{figure}

For the depth image, a more thorough process is executed. Let $D$ be the pre-processed depth image. $n$ new binary images $D_1, D_2, ... , D_n$, having the same size as $D$, are created. Each image represents a layer in the depth dimension (Fig. \ref{img_layers}), with the layers separated from each other by $t/n$ cm. The pixels in the layers are assigned values according to Eq. \ref{eqn_succdept}, where $i$ is a pixel and $l, 1 \leq l \leq n$ is the layer.

\begin{equation}
 D_l(i,l) = \begin{cases}
        1  & \text{if} \; D(i) \leq ((l-1) \times \cfrac{t}{n}) + 1 \\
        0 & \text{otherwise}
        \end{cases}
\label{eqn_succdept}
\end{equation}

\begin{figure}
 \centering
 \includegraphics[width=0.5\textwidth]{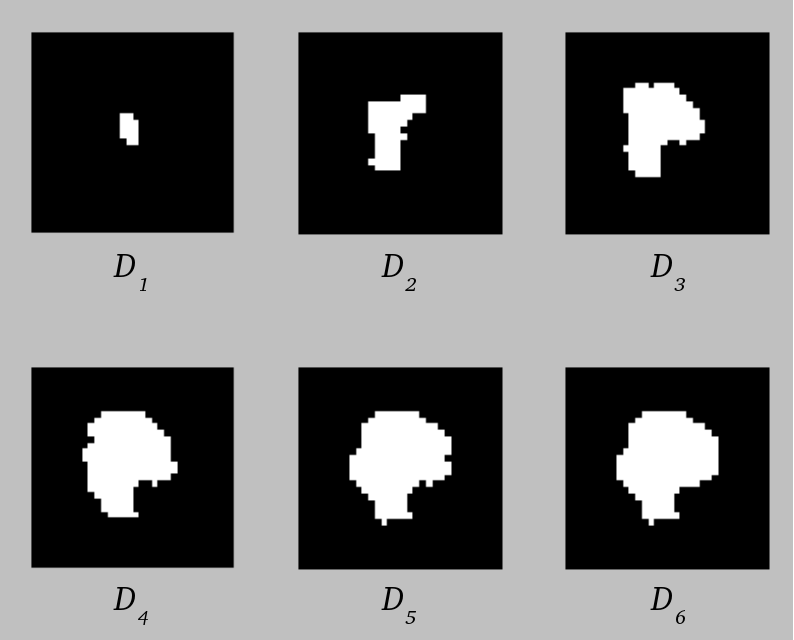}
 \caption{Successive binary depth images for a hand doing the ``G'' sign. $D_1$ is the tip of the two forward-pointing fingers, $D_2$ and $D_3$ add more fingers, $D_4$ includes the hand except for a part of the palm, and more parts of the palm can be seen on $D_5$ and $D_6$.  Here, we used $n=6$ layers.}
 \label{img_succdpt}
\end{figure}

Finally, each layer is centered using the bounding box technique. Fig. \ref{img_succdpt} illustrates this process.

In our experiments, we found that using $n=6$ binary images and a maximum hand depth of $t=12$ cm gave the best features, so that the layers $D_l$ are separated by 2 cm. This distance is consistent with the approximate thickness of a finger, and being able to differentiate finger position is key in distinguishing one sign from another. Also, the fingers are always located on the front part of the hand, well behind our 12 cm margin.

To generate the depth feature vector $f_{depth}$, each of the $n$ binary depth images is first resized to 32x32, unrolled into a 1x1024 vector and all of them are concatenated in a 1x6144 vector. Finally, the intensity image features $f_{intensity}$ and depth image features $f_{depth}$ are concatenated together into a feature vector $f_{combined}$ of size 1x10240.

\subsection{Deep Belief Network}

For our classifier, we used a DBN composed of three Restricted Boltzmann Machines (RBM) and one additional translation layer. DBNs are similar to MLPs, but they have many more layers. The additional layers provide more learning potential, yet they are much harder to train \cite{hinton2006reducing}. Nonetheless, recent work has made training them possible and they have a very promising future. Much of contemporary machine learning research focuses on deep learning, including DBNs. 

Instead of using classical MLP learning algorithms such as backpropagation, the hidden layers are first trained individually as RBMs. Starting with the data, the activations of each layer are successively fed to the next RBM as training data. When all layers are trained, they are all stacked as an MLP and an additional translation layer is added to the top.

Each RBM layer is trained using Constrastive Divergence-1 (CD-1), which is an unsupervised algorithm to make the network learn patterns in the data. This algorithm is especially well suited with binary units, which is the reason we extracted binary features from the data via the depth images $D_l$. Other types of inputs can be used, but they either require more units or a much smaller learning rate, which makes the network harder and slower to train \cite{hinton2012practical}.

\section{Experiments}

Before experimenting with our new feature extraction technique, we looked at other ways to extract features from the depth and intensity images. As a baseline experiment, we trained a DBN on the raw intensity and depth images. For this experiment, the images were pre-processed as explained in Sec. \ref{sec_features}. We kept the original size of the pictures (128x128), then unrolled them into two 1x16384 vectors with which the DBN was trained.

A second experiment was done using Gabor filters on the processed images, using four different scales and orientations. The parameters were fine-tuned by hand in order to extract the key points of the images that would be useful in the classification step. Due to the large size of the features produced by our Gabor filters on 128x128 images, we had to resize their output to 28x28 in order to make training possible. This small size made the features very limited since much of the information was lost during the resizing.

We also tried to apply bar filters to our depth and intensity images in order to extract the main contours of the hands. Three filters were used; one horizontal, one diagonal and one vertical. This reduced number of filters (3 instead of 16 for the Gabor filters) made it possible to keep their activations to larger sizes. However, many details were still lost since most of the hand contours are not precisely aligned to these three orientations, and doing so made some signs more difficult to distinguish by the classifier.

We tested our technique on the same data set as \cite{pugeault2011spelling}. This data set contains all 24 static letters of the fingerspelling alphabet, signed by five different people. There are about 500 images of each letter for each person, resulting in over 60000 total examples. The images feature many different viewpoints, as participants were asked to move their hands around while doing each sign. Some of the images have the hands nearly sideways, some others show the front, others are from a higher or lower perspective. Images were captured using the Microsoft Kinect sensor and the data set includes, for each image, an intensity and a depth version. However, the pixels in the intensity images do not map to the same location in the depth images in this data set. To correct this, we had to scale and translate the intensity images to match the depth images.

Two different approaches to generating the training and testing data sets were tried in order to compare with previous work reported in the literature. The first, $T_{allseen}$, was to split the data set in a training set, a validation set and a test set. The training set contained half of the examples, and the validation and test sets each contained a quarter of the examples. The split was done in such a way that all five people are equally represented in all sets.

The other technique, $T_{unseen}$, consisted of using the images of four people for the training and validation sets. Around five thousands images (representing $1 / 10$ of the dataset size) was used for the validation set, while the rest was kept for training. The images of the fifth person were shuffled and used as test set. Each of the five people were used as test sets against the four remaining people and we averaged the results to avoid any bias.

\begin{figure}
 \centering
 \includegraphics[width=0.5\textwidth]{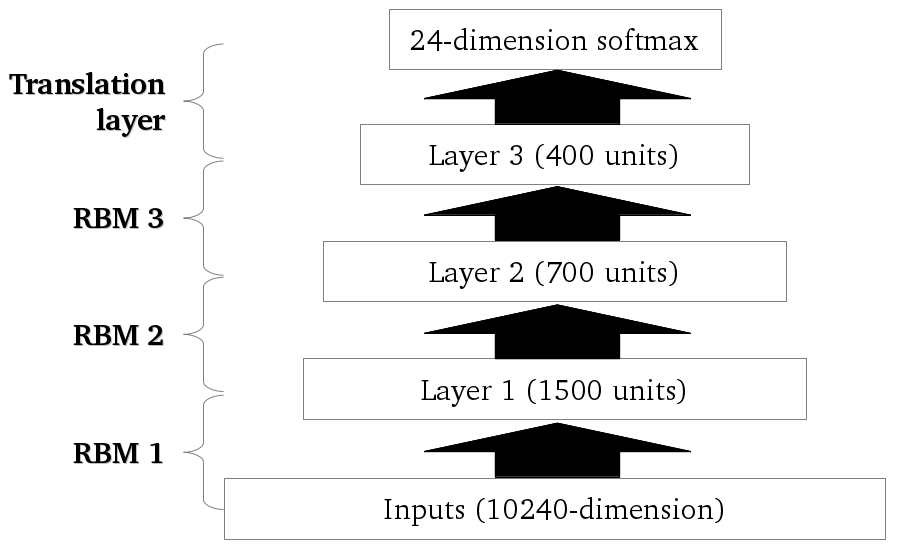}
 \caption{Representation of the DBN we used. The feature vector $f_{combined}$ is the input, and the output indicates which of the 24 static letters has been identified.}
 \label{fig_dbn}
\end{figure}

We found that a three layer network, using the new feature extraction technique, worked best for this task, with respectively 1500, 700 and 400 units on each layer (see Fig. \ref{fig_dbn}). We trained our network in three steps. Each hidden layer was successively trained with CD-1 for 60 epochs, or until convergence. We used both L1 and L2 weight decay to avoid overfitting.

The second step was to train the translation (output) layer. We trained this layer using regular backpropagation to translate the activations of the different RBMs into a 24 dimension softmax for each of the 24 letters. In this step, we only changed the weights of this translation layer, in order to predict the class from what the RBMs had already learned. Since this layer was initialized randomly, modifying the weights of the pre-learned layers would be counter-productive as it would use the gradient of random weights as an error function to the RBM layers. We did 200 epochs of backpropagation, and used both L2 weight decay and early stopping. The validation set was used to decide when to stop learning and to adjust the learning metaparameters. Additionnally, we added gaussian noise in the inputs as explained in \cite{da2011pca}. Our algorithm also made use of momentum.

The last step was a fine backpropagation step, using the whole network this time, but with a much lower learning rate. We also made use of the same metaparameters as the last step, reducing some values to compensate for the lower learning rate.

\section{Results}

\begin{figure}
 \centering
 \includegraphics[width=0.5\textwidth]{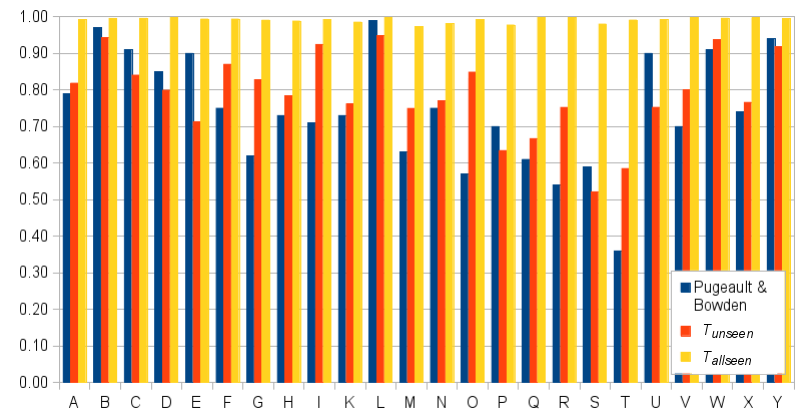}
 \caption{Comparison of the precision for all letters between our two testing methods and \cite{pugeault2011spelling}. Pugeault \& Bowden use $T_{allseen}$.}
 \label{chart_precision}
\end{figure}

\begin{figure}
 \centering
 \includegraphics[width=0.5\textwidth]{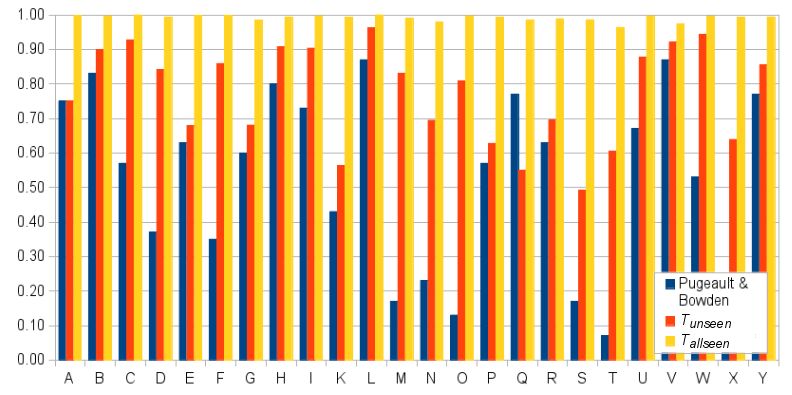}
 \caption{Comparison of the recall for all letters between our two testing methods and \cite{pugeault2011spelling}. Pugeault \& Bowden use $T_{allseen}$.}
 \label{chart_recall}
\end{figure}

We evaluated our results with precision and recall measures, computed for each letter. We compared both testing approaches with the results of Pugeault \& Bowden \cite{pugeault2011spelling} as they use the same data set. A detailed comparison of our results is presented in Fig. \ref{chart_precision} for the precision measure and Fig. \ref{chart_recall} for the recall measure.

We first evaluated our three initial attempts at feature extraction using the $T_{unseen}$ method. For the raw intensity and depth vectors, we obtained 68 \% recall and 73 \% precision. For the Gabor filters, we obtained 64~\% recall and 69~\% precision, and for the bar filters, 67~\% recall and 71~\% precision. These results were obtained with the same DBN as our other results. We can see than the DBN was able to learn some interesting features from the raw images, which gave better results than the Gabor and bar filters.

We then compared those results with the layer-based feature extraction combined with the intensity image feature extraction we described in Sec. \ref{sec_features}. As the figures show, we obtain 99~\% recall and precision for the $T_{allseen}$ testing approach, in comparison to 53~\% recall and 75~\% precision for Pugeault \& Bowden who use the same $T_{allseen}$ approach \cite{pugeault2011spelling}. Thus, we believe that our feature extraction technique gave us useful features and that our deep neural network learned a good internal representation of the signs. 

\begin{figure}
 \centering
 \includegraphics[width=0.5\textwidth]{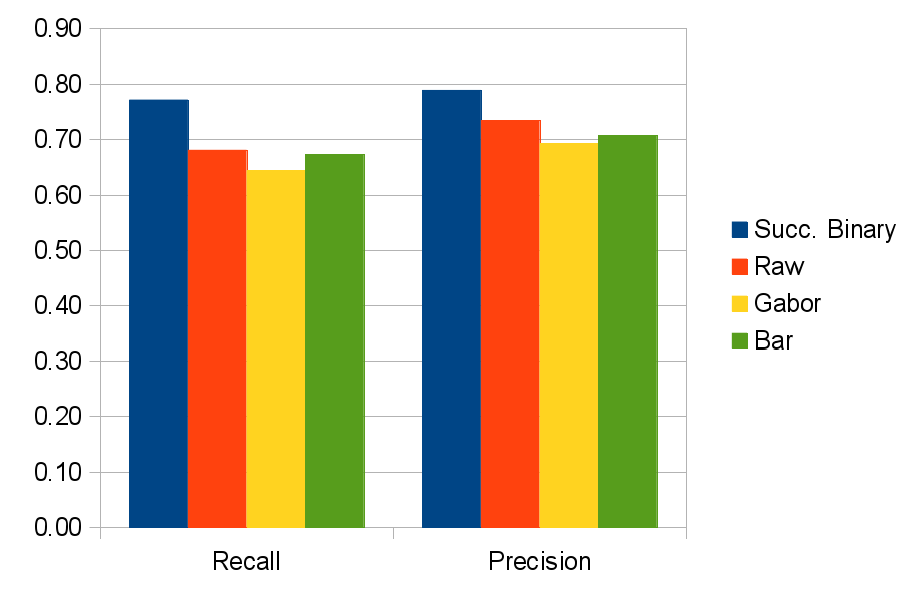}
 \caption{Comparison of the recall and precision for the different types of features used in the $T_{unseen}$ method.}
 \label{chart_comparison}
\end{figure}

For the other testing approach $T_{unseen}$, which is harder than $T_{allseen}$ since a new user is presented to the network, we achieve 77~\% recall and 79~\% precision. This is lower than our results for $T_{allseen}$, but is to be expected since many signs are very similar and that different signers have different ways of making the signs as explained in Sec. \ref{sec_intro}. This is a more realistic test, for the cases where a recognition system is trained beforehand, but used in a system with different users. We can also see that those results are higher than all of the previous tests we did with the Gabor, raw images and bar filters (Fig. \ref{chart_comparison}).

The most confused letters are E (15~\% recall) with S, Q (23~\% recall) with P, K (49~\% recall) with V and N (50~\% recall) with T. All of these signs differ very little from each other, making it difficult for a learning algorithm to determine which letter the sign really was with a new, unseen user. 
Note that the use of bigram, prior distribution on words and dictionary would alleviate these issues (as used by  \cite{kim2012american}), but would not be able to correct for things such as passwords or letter-based codes.

However, our system was able to distinguish with good accuracy the letters M and N (83 \% and 69 \% recall respectively), with little confusion (under 10 \% between them). Those signs are very hard to distinguish from each other using only the intensity data (see Fig. \ref{fig_SimilarGestures}), but this task becomes easier using the depth data. This shows that the layer-based feature extraction technique provides non-negligible information for classifying hand signs in 3D.

\section{Conclusion and Future Work}

We have presented a novel way to extract features from depth images along with a way to segment the hand on intensity images using depth images. We have combined this technique with a Deep Belief Network for the task of ASL fingerspelling classification. We demonstrated the value of these new features by comparing them against more classical features (raw images, Gabor and bar-like). We obtained excellent results on two different scenarios, including one with a new unseen user.

However, some restrictions remain. Our system is only capable of classifying static gestures. However, ASL fingerspelling is not only static, but also includes two dynamic gestures. Many other applications could also benefit from this extended capability. Work could also be done to allow the usage of both hands instead of one, which could make comparison possible on other approaches for different sign languages, such as British Sign Language that is two handed.

Also, our system is limited by the staticity of the dataset we used. In a practical fingerspelling environment, users make signs one after the other. Thus, they have to change the position of their fingers and hand from one sign to the other, keeping the final shape only for a very brief moment. In comparison, our system takes as input an image of a sign, taken when the hand is movementless for the duration of the pose. This is easier since all the transition phase is already filtered out. However, this could be corrected by forcing the user to keep the pose for a few seconds, and detecting those static movements in our algorithm.

{\small
\bibliographystyle{ieee}
\bibliography{biblio}
}
\end{document}